\documentclass[10pt,twocolumn,letterpaper]{article}

\usepackage{iccv}
\usepackage{times}
\usepackage{epsfig}
\usepackage{graphicx}
\usepackage{amsmath}
\usepackage{amssymb}

\usepackage[breaklinks=true,bookmarks=false]{hyperref}
\usepackage{multirow}
\usepackage{graphicx}
\usepackage{enumitem}
\usepackage{amsmath}
\usepackage{amssymb}
\usepackage{booktabs}
\usepackage{verbatim}
\usepackage{tabularx,booktabs}
\newcolumntype{Y}{>{\centering\arraybackslash}X}
\usepackage{graphicx}
\usepackage{bbding}
\usepackage{subcaption}
\usepackage{xcolor}
\usepackage[accsupp]{axessibility}  
\usepackage{hyperref}
\hypersetup{
    colorlinks=true,    
    urlcolor=red,
 }
 \usepackage{cleveref}
 \usepackage[toc,page]{appendix}

\graphicspath{ {./data/} }

\iccvfinalcopy 

\def\iccvPaperID{****} 

\ificcvfinal\fi

\begin{document}

\title{RSFNet: A White-Box Image Retouching Approach using Region-Specific Color Filters}

\author{
Wenqi Ouyang$^{1}$,
Yi Dong$^{2}$,
Xiaoyang Kang$^{1}$,
Peiran Ren$^{1}$,
Xin Xu$^{1}$,
Xuansong Xie$^{1}$
\\
$^{1}$DAMO Academy, Alibaba Group,
$^{2}$Nanyang Technological University
\\
{\tt\small wenqi.oywq@alibaba-inc.com,}
{\tt\small ydong004@ntu.edu.sg,}\\
{\tt\small \{peiran.rpr, kangxiaoyang.kxy, chris.xx, xingtong.xxs\}@alibaba-inc.com}
}

\maketitle
\ificcvfinal\thispagestyle{empty}\fi

\begin{abstract}
Retouching images is an essential aspect of enhancing the visual appeal of photos. Although users often share common aesthetic preferences, their retouching methods may vary based on their individual preferences. Therefore, there is a need for white-box approaches that produce satisfying results and enable users to conveniently edit their images simultaneously. Recent white-box retouching methods rely on cascaded global filters that provide image-level filter arguments but cannot perform fine-grained retouching.
In contrast, colorists typically employ a divide-and-conquer approach, performing a series of region-specific fine-grained enhancements when using traditional tools like Davinci Resolve. We draw on this insight to develop a white-box framework for photo retouching using parallel region-specific filters, called RSFNet. Our model generates filter arguments (e.g., saturation, contrast, hue) and attention maps of regions for each filter simultaneously.
Instead of cascading filters, RSFNet employs linear summations of filters, allowing for a more diverse range of filter classes that can be trained more easily. Our experiments demonstrate that RSFNet achieves state-of-the-art results, offering satisfying aesthetic appeal and increased user convenience for editable white-box retouching. Code is available at {\small \url{https://github.com/Vicky0522/RSFNet}}.\end{abstract}
\begin{figure}
    \centering
    \includegraphics[width=0.45\textwidth]{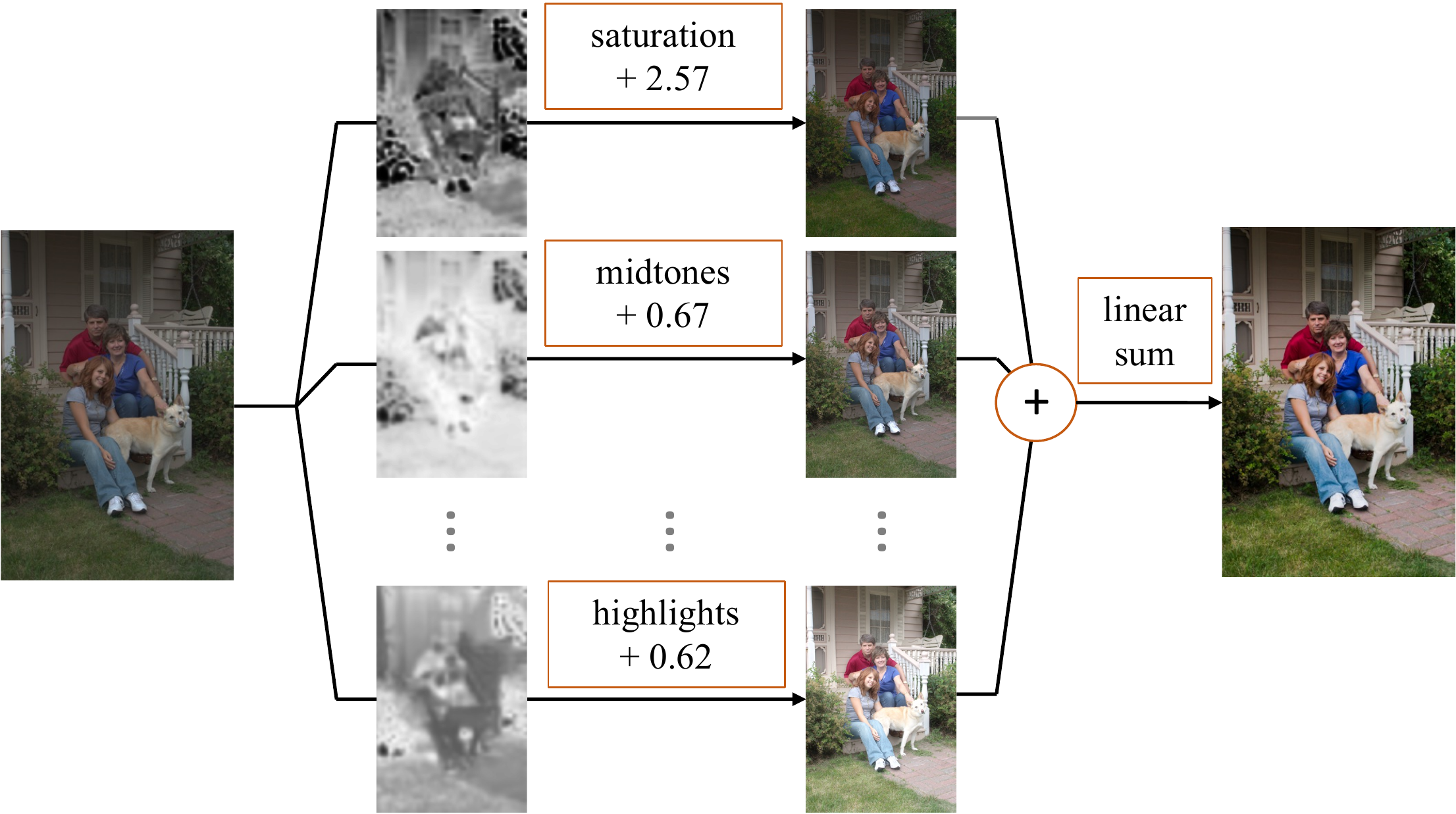}
    \caption{Architecture of our white-box retouching framework. Our model generates filter arguments (e.g., saturation, midtones, highlights) and attention maps of regions for corrsponding filter simultaneously. Final result is obtained by conducting linear summations on filtered results. }
    \label{fig:img_model_1}
\end{figure}

\section{Introduction}
\label{sec:1}

Photos and videos recorded by the camera usually lack aesthetic quality due to poor shooting condition and inexperienced photographer. Artists often use professional-grade softwares (\eg, PhotoShop for image, Davinci Resolve for video) to enhance image and video quality. However, it requires professional retouching skills to conduct a series of sophisticated manual adjustments. The use of fool-proof applications that present various style templates simplifies the retouching procedure, but it is unable to achieve optimal results due to the lack of enhancement capability. 

Recent learning-based methods have demonstrated the strong capability of deep neural nets for automatic photo retouching. Automatic systems are established to generate the optimal result end-to-end. However, one of the most significant considerations is that retouching is not a problem with an exclusive solution. People have different retouching preferences. Even the same artist may retouch the same image in different styles to meet various demands. In order to provide convenience for manual edits, automatic retouching systems must provide not only the suggested result, but also the retouching strategy in a way understandable by humans. 

Taking these considerations into account, we propose a white-box framework that uses the divide-and-conquer strategy employed by artists in traditional retouching tools. Our model generates attention maps of regions as well as filter arguments for traditional edits for these regions. This allows users to alter the suggested results to their preferences.

The main {\bf contributions} of this work are as follows:

 \begin{itemize}[itemsep=0pt,topsep=0pt,parsep=0pt]
 \item We redefine the retouching problem according to the divide-and-conquer strategy, focusing on finding attention maps for regions and human-understandable filter adjustments per map to achieve the best results.
 
 \item We propose RSFNet, which generates pixel-level attention maps of regions and filter arguments simultaneously. By conducting linear summations on the filtered results, our model demonstrates superior performance to global white-box retouching methods, with a wider range of filter classes for fine-grained enhancement and a simpler training procedure.
 
  \item We propose a scheme that allows users to edit RSFNet suggested results, demonstrating its effectiveness and convenience in image applications.
\end{itemize}

\section{Related Works}
\label{sec:2}

{\bf Deep Image Enhancement.}
Many attempts have been made in image enhancement using learning-based methods. These methods can broadly be classified into two categories: deep-convolutional-based models and physical-inspired models. The former~\cite{DBLP:conf/cvpr/ChenCXK18,DBLP:conf/cvpr/ChenWKC18,DBLP:conf/mm/DengLT18,DBLP:conf/bmvc/WeiWY018,DBLP:conf/mm/ZhangZG19} view enhancement as a synthesis problem and use fully convolutional generators~\cite{DBLP:conf/cvpr/LongSD15} to achieve dense image-to-image translation. While these methods exhibit strong capability for generating and enhancing details, they suffer from heavy structure and low inference speed. The latter category of works defines enhancement as the parameter prediction of a physical model and uses deep learning strategies to fit the model. These models include 3D-LUT~\cite{DBLP:conf/iccv/Wang0PMWSY21,zeng2020lut,yang2022adaint}, parametric filters~\cite{DBLP:conf/cvpr/MoranMMPS20}, conditional sequential global modulation~\cite{DBLP:conf/eccv/HeLQD20}, affine color transformations~\cite{DBLP:journals/tog/GharbiCBHD17,DBLP:conf/cvpr/WangZFSZJ19}, 1D mapping curves~\cite{DBLP:conf/iccv/Song0D21,Kim_2021_ICCV,DBLP:conf/icpr/MoranMS20}. Their enhancement capabilities depend on the range of transformation functions the physical model covers. Among those, models based on global 3D-LUT~\cite{yang2022adaint,zeng2020lut} are well-designed for retouching and could achieve high performance with fast inference speed. However, they suffer from uneven transitions in smooth areas due to lack of fine-grained local adjustment. Spatial-Aware 3D-LUT~\cite{DBLP:conf/iccv/Wang0PMWSY21} alleviate the problem of global 3D-LUT by computing pixel-level weight maps. All these methods have a black-box structure or unintuitive parameters that are hard for humans to understand. DeepLPF~\cite{DBLP:conf/cvpr/MoranMMPS20} has been a significant milestone in the field of region-specific color enhancement via local parametric filters but suffers from low inference speed and limited filter shapes. The three types of filters with three implementations each results in complex maps that are hard to understand.

{\bf White-box Image Editing.}
Recent white-box methods~\cite{hu2018exposure,Harmonizer,DBLP:journals/tog/YanZWPY16,DBLP:conf/cvpr/ZouSQ0S21} have decoupled image editing into a series of human understandable operations and deep learning strategies to predict them. Of these, our work is most related to~\cite{Harmonizer,hu2018exposure}. However, they train networks with cascaded filter argument prediction modules to perform global retouching step-by-step. Our model utilizes pixel-wise region-specific filters to capture local features and employs linear summations to combine filters, thus having stronger capability to cover a wider range of color transformation functions.

\section{Method}
\label{sec:3}

\subsection{Design Motivation}
\label{sec:3.1}
{\bf Traditional Retouching Strategy.}
In the traditional retouching process, local adjustments for different regions are conducted separately and then aggregated together to accomplish fine-grained enhancement. All the adjustments could be accomplished in one {\em Layer}, where filters are all conducted on the same image instead of results from previous filters. We adopt this divide-and-conquer strategy to build our framework. We select $10$ commonly used retouching filters from traditional tools(\eg, Davinci Resolve) to represent adjustment manipulations, including {\em contrast, saturation, hue, temperature, shadows, midtones, highlights} and {\em shift}. For more details about filters, please refer to Appendix \ref{appendix:a} . Therefore, retouching is defined as finding pixel-level attention maps and corresponding adjustments for each map to achieve the most desirable result. We adopt this convention and represent retouching result $Y$ as adding linear summations of increments resulted by filters to the original image as: 

\begin{equation}
\begin{aligned}
  Y =  X + \sum_{m} \sum_{n} (F_{m,n}(\theta_{m,n}, X)-X)\odot M_{m}
  \end{aligned}
  \label{eq:3.1_1}
\end{equation}

Where $\theta_{m,n}$ and $F_{m,n} \in \{contrast, ..., shift\}$ are the argument value and filter function of the $n$th filter for the $m$th attention map of input image $X$ respectively, $M_{m}$ is the $m$th attention map for image $X$. It should be noticed that this linear representation makes our optimization process much simpler than other white-box methods.

\subsection{Architecture}
\begin{figure*}
    \centering
    \includegraphics[width=0.9\textwidth]{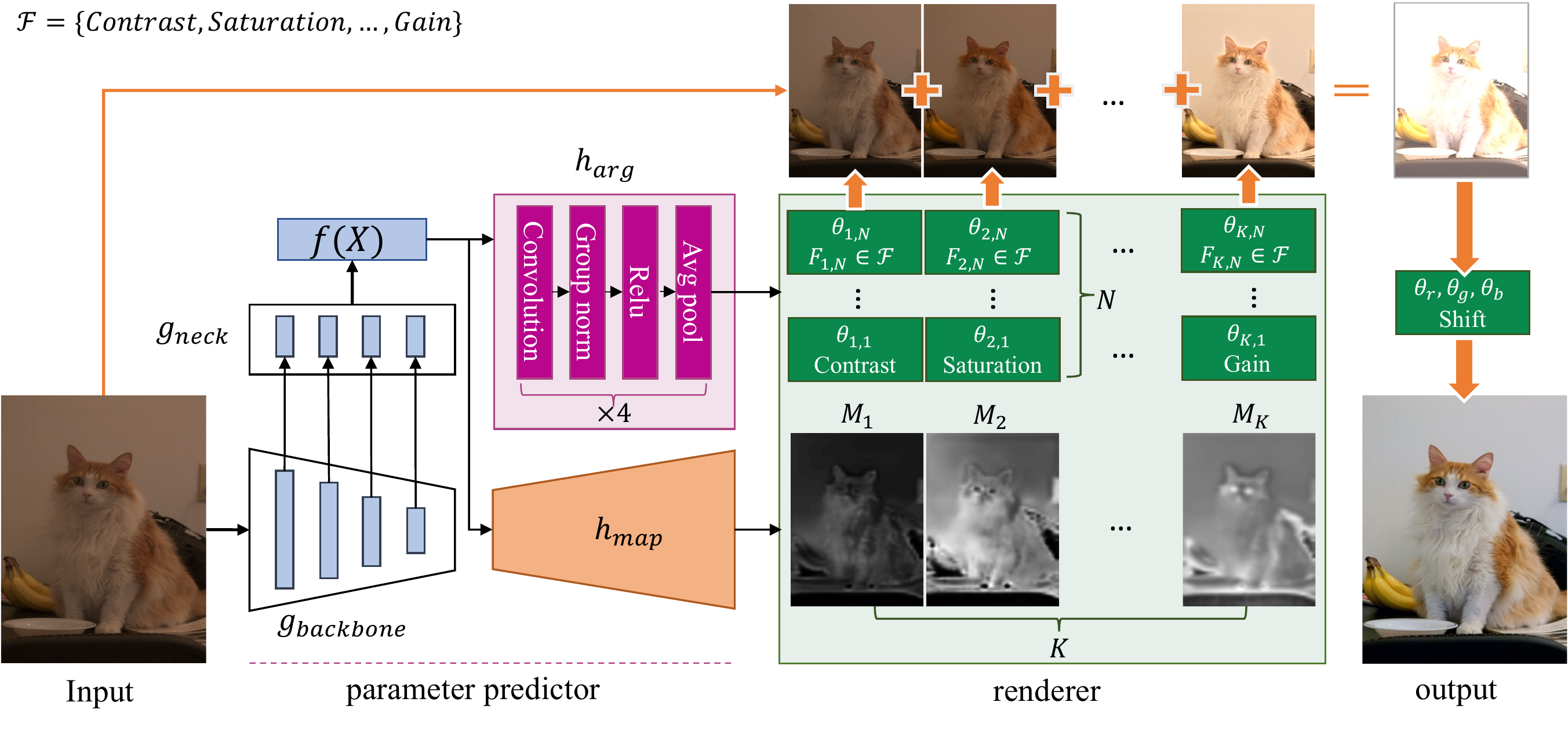}
    \caption{Our network comprises two modules: the parameter predictor and renderer. The input image is transformed into feature embeddings $f(X)$ through a backbone $g_{backbone}$ and FPN neck $g_{neck}$, which are then utilized by the map generator $h_{map}$ and argument regressor $h_{arg}$ to generate region maps and filter arguments (e.g. contrast, hue, saturation), respectively. The renderer applies the filter arguments to each region map and aggregates them to produce the final retouching result.}
    \label{fig:3.2_1}
\end{figure*}
\label{sec:3.2}
Our model consists of two modules, the parameter predictor and the renderer. The architecture of the parameter predictor shares a similar overall structure with previous segmentation models~\cite{wang2020solo,wang2020solov2}. As illustrated in Figure~\ref{fig:3.2_1}, the network $g$ is composed of a backbone $g_{backbone}$, a FPN neck $g_{neck}$, a map generator $h_{map}$ for attention map prediction and an argument regressor $h_{arg}$ for arguments regression. We set the last layer of map generator $h_{map}$ to sigmoid, followed by an upsampling layer to match the size of the original input image. We use $f(X)$ to represent the output features of FPN neck. Maps generated by $h_{map}$ after upsampling are: 
\begin{equation}
\begin{aligned}
  M_{m} =  h_{map}(f(X))_{m}, m \in \{1,....,K\} \\
  \end{aligned}
  \label{eq:3.2_1}
\end{equation}
 $K$ is the number of output channels of the last layer. The argument regressor $h_{arg}$ consists of $4$ units of convolutional modules. Each module consists of a convolution layer followed by group normalization, $relu$ activation and average pooling. Output feature $f(X)$ is finally reduced by $h_{arg}$ to a $K \times{N}$ vector representing arguments of all filters for every map:
 \begin{equation}
\begin{aligned}
    \theta_{m,n} =  h_{arg}(f(X))_{m,n}, m \in \{1,....,K\}, n \in \{1,...,N\} \\
  \end{aligned}
  \label{eq:3.2_2}
\end{equation}
$K$ is the number of color filters per map. Given the maps and arguments from the parameter predictor, the renderer applies filter functions $F_{m,n} \in \{contrast, ..., shift\}$ to the input image and merges filtered results via linear summations of increments. The final output image $Y$ in Equation~\ref{eq:3.1_1} is:
\begin{equation}
 \begin{aligned}
  Y = X + \sum_{m=1}^{K} \sum_{n=1}^{N} (F_{m,n}(\theta_{m,n}, X)-X) \odot M_{m}
  \end{aligned}
  \label{eq:3.2_3}
\end{equation}
In our implementation, we set the number of filters per map as $N=1$, indicating that there is one specific filter assigned to each map. This is illustrated in the second row of filter boxes depicted in Figure~\ref{fig:3.2_1}. We also consider the filter {\em Shift} as a global function that is applied to the entire image without an attention map. Through our experimentation, we observe that this configuration is sufficient for generating satisfactory results while also providing convenience for users to edit.

{\bf Training Loss.}
The overall framework can be trained end-to-end. Our training loss function is defined as follows:
\begin{equation}
 \begin{aligned}
  L=L_{recon},
   \end{aligned}
\label{eq:3.2_4}
\end{equation}
where $L_{recon}$ is the $l1$ loss for reconstruction. \

\subsection{A Variation of RSFNet with Controlled Region Shape}
\begin{figure}
    \centering
    \includegraphics[width=0.47\textwidth]{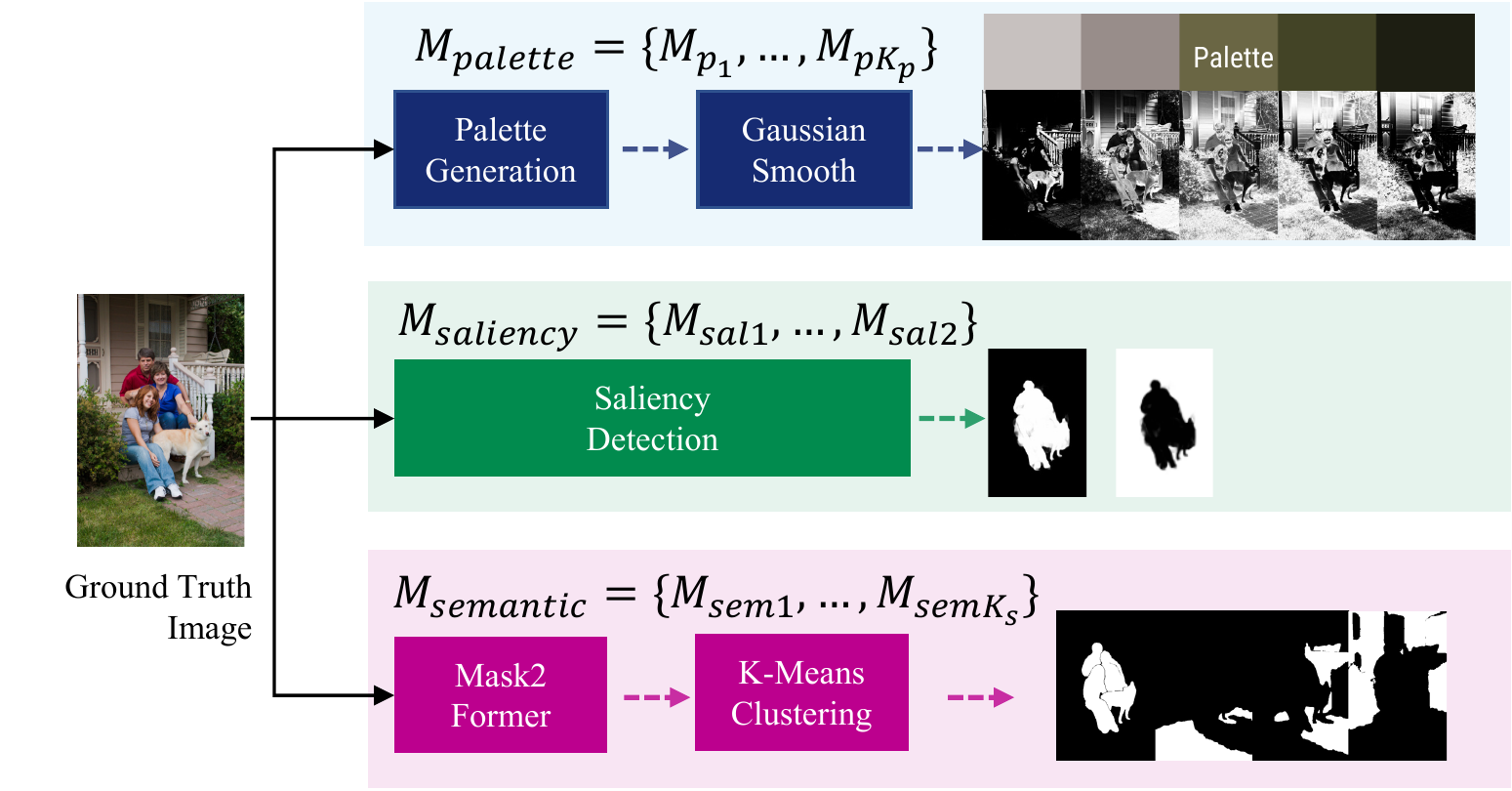}
    \caption{A variation of RSFNet could be used to generate masks with controlled shapes. We generate ground truth masks for training via three types of region masks derived from off-the-shelf models: palette-based segmentation, semantic segmentation, and saliency detection.}
    \label{fig:3.3_1}
\end{figure}
\label{sec:3.3}
\begin{figure*}[t]
  \centering
   \includegraphics[width=0.99\linewidth]{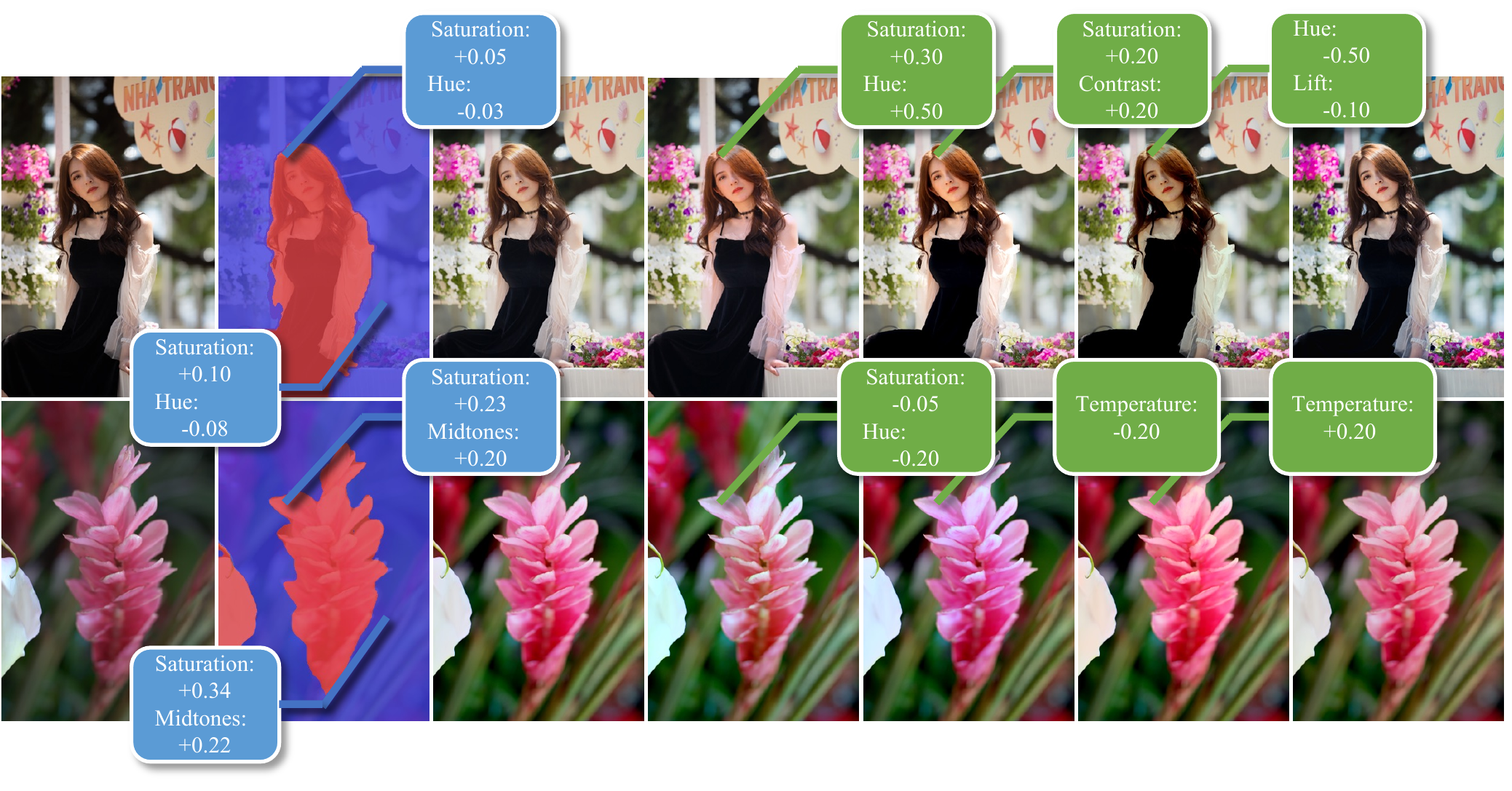}
   \resizebox{\textwidth}{!}{%
  \footnotesize
   \begin{tabular}{p{2.05cm}<{\centering} p{2.05cm}<{\centering} p{2.05cm}<{\centering} p{2.05cm}<{\centering} p{2.05cm}<{\centering} p{2.05cm}<{\centering} p{2.05cm}<{\centering}}
input & Result(Args \& Masks) & Result(Image) & Edit Version 1. & Edit Version 2. & Edit Version 3. & GroundTruths
\end{tabular}%
}
   \caption{Editable white-box retouching. The second column illustrates the arguments and masks generated by RSFNet-saliency that is trained with saliency masks. The third column displays the retouched result. On the right-hand side, the figure showcases three versions of adjustments conducted on the retouched results, with numbers denoted in green boxes indicating the relative incremental value of filters.}
   \label{fig:img_edit}
\end{figure*}

Our proposed framework has the capability to generate region masks with controlled shapes, with minimal modifications to the training data and loss function. In our observation, artists tend to select regions based on their main color, saliency, or other high-level semantic information. To leverage this insight, we utilize palette-based methods~\cite{Chang_2015_PPR}, saliency detection~\cite{liu2019simple}, and panoptic segmentation networks~\cite{cheng2021mask2former} to generate palette-based, saliency, and semantic masks of the input images used for training, as depicted in Figure~\ref{fig:3.3_1}. Further implementation details can be found in Appendix \ref{appendix:b}.
Our training loss function in Equation~\ref{eq:3.2_4} is modified as follows:
\begin{equation}
 \begin{aligned}
  L=L_{recon} + \lambda L_{mask},
 \end{aligned}
\label{eq:3.2_5}
\end{equation}
where $L_{recon}$ is the $l1$ loss for reconstruction. When training with palette-based masks and saliency masks, $L_{mask}$ is the Dice Loss for mask prediction following loss functions in~\cite{wang2020solo,wang2020solov2}. We change it to the Cross Entropy Loss when training with semantic masks, at the time map generator $h_{map}$ ends with a softmax layer.

The model generates region masks and corresponding filter arguments for the input image simultaneously. Users can select region maps and adjust filter arguments to edit results. We show an example of {\em RSFNet-saliency} trained with saliency masks in Figure~\ref{fig:img_edit}.
\begin{table*}[]
\footnotesize
\begin{tabular}{@{}ccc@{}}
\begin{tabular}[c]{@{}c@{}}
\begin{subtable}[t]{0.33\textwidth}
\begin{tabular}{@{}cc|cc@{}}
\toprule
\begin{tabular}[c]{@{}c@{}}mask as\\ input\end{tabular} & mask type     & PSNR$\uparrow$ & SSIM$\uparrow$ \\ \midrule
w/                                                     & saliency(1)   & 23.83    & 0.906   \\
w/                                                     & palette(5)    & 23.45    & 0.903  \\
w/                                                    & semantic(10)  & 21.65    & 0.866   \\
w/                                                     & semantic(5)   & 21.77    & 0.867   \\
 w/o                                                    & saliency(1)   & 24.01    & {\bf0.909}    \\
 w/o                                                     & palette(5)    & 24.14    & 0.906    \\
 w/o                                                    & semantic(10)  & 24.09    & 0.908    \\
 w/o                                                    & semantic(5)   & {\bf24.17}    & {\bf0.909}    \\
  w/o                                                   & semantic(133) & 23.75    & 0.905    \\ 
\bottomrule
\end{tabular}
\caption{Analysis on using mask as input and different mask types. Bracket after ``mask type" indicates number of masks.}
\label{tab:4.3_1.1}
\end{subtable} \end{tabular} &
\begin{tabular}[c]{@{}c@{}}
\begin{subtable}[t]{0.33\textwidth}
\centering
\begin{tabular}{@{}cc|cc@{}}
\toprule
\begin{tabular}[c]{@{}c@{}}mask\\ first\end{tabular} & masktype              & PSNR$\uparrow$ & SSIM$\uparrow$                 \\ \midrule
w/ & saliency(1) & 23.26 & 0.890 \\
w/o  & saliency(1) & {\bf 24.01} & {\bf 0.909} \\
\multicolumn{1}{l}{}                                 & \multicolumn{1}{l|}{} & \multicolumn{1}{l}{} & \multicolumn{1}{l}{} \\ \bottomrule
\end{tabular}
\caption{Analysis of training order. ``Mask first" refers to building model on pre-trained PoolNet~\cite{liu2019simple}.}
\label{tab:4.3_1.2}
\end{subtable}
\end{tabular} &
\begin{tabular}[c]{@{}c@{}}
\begin{subtable}[t]{0.33\textwidth}
\centering
\begin{tabular}{@{}cc|cc@{}}
\toprule
backbone & \begin{tabular}[c]{@{}c@{}}downscaling\\ factor\end{tabular} & PSNR$\uparrow$ & Runtime$\downarrow$ \\ \midrule
resnet18 & x8 & 24.65 & 9.98 \\
resnet18 & x4 & {\bf24.83} & 12.38 \\
resnet10 & x8 & 24.50 & {\bf7.98} \\
resnet10 & x4 & 24.38 &  8.71 \\ \bottomrule
\end{tabular}
\caption{Trade off between speed and accuracy.}
\label{tab:4.3_1.3}
\end{subtable}
\end{tabular}\end{tabular}
\caption{Ablation analysis of our framework. Models are trained on the FiveK dataset~\cite{fivek} with inputs zeroed as shot.}
\label{tab:4.3_1}
\end{table*}

\section{Experiments}
\label{sec:4}
\subsection{Datasets and Application Settings}
\label{sec:4.1}
We conduct experiments on two publicly available datasets: MIT-Adobe FiveK~\cite{fivek} and PPR10K~\cite{ppr10k}. The MIT-Adobe FiveK consists of 5,000 RAW images with their retouched versions. We follow prior works~\cite{DBLP:conf/eccv/HeLQD20,Kim_2021_ICCV,zeng2020lut,yang2022adaint} to adopt images retouched by expert C as ground truths and split the dataset into 4,500 pairs for training and 500 pairs for validation. Images are resized to 480p during training stage, whereas both of 480p resolution and original resolution are used during validation. We follow the official split~\cite{ppr10k} to split PPR10K dataset into 8,875 pairs for training and 2,286 pairs for testing. Images are resized to 360p resolution for training and validation. 

For the MIT-Adobe FiveK dataset, we conduct experiments on two input settings: {\em input zeroed with expertC white balance} and {\em input zeroed as shot}. The second set of inputs is more challenging as it also deteriorates in white balance. Therefore, to restore its visual appeal, our framework's ability to perform color temperature correction and hue adjustment is required.

\begin{figure}
    \centering
   \includegraphics[width=0.45\textwidth]{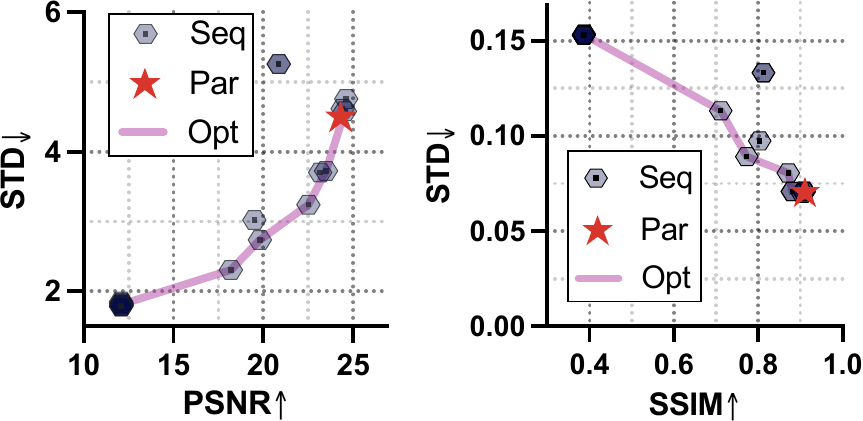}
    \caption{Comparison between sequential and parallel methods under the same filter functions and network structures.}
    \label{fig:seq_par}
\end{figure}

\subsection{Implementation Details}
\label{sec:4.2}
To implement our proposed framework as described in Section~\ref{sec:3.2}, we employ ResNet18~\cite{resnet} as the backbone and utilize the output features of layer $(0,1,2,3)$ as concatenated input with $256 \times{4}$ channels for the FPN neck. The FPN neck reduces the features to $64 \times{4}$ channels and feeds them into the map generator and argument regressor. We set the output numbers $K$ of region maps to $10$ and $16$ for the Adobe-MIT FiveK and PPR10K datasets, respectively. The number of filters per map $N$ is set to $N=1$, with $10$ filters used for MIT-Adobe FiveK, $1$ each for {\em shadow, midtones, highlights}, where arguments for RGB channels are set equal. For PPR10K, we use $16$ filters, with $3$ each for {\em shadow, midtones, highlights}, where arguments for RGB channels are set differently.

To implement the variation of RSFNet with controlled region shapes, as described in Section~\ref{sec:3.3}, we generate ground truth masks for the MIT-Adobe FiveK dataset. Specifically, we generate five palette-based masks, one saliency mask, and five semantic masks for each image. For the PPR10K dataset, we directly use the human-region masks provided in the dataset as ground truth masks for training.
The map generator is designed to output $5$ channels for the palette-based version, $2$ channels for the saliency version, and $5$ channels for the semantic version. We use $10$ filters for the MIT-Adobe FiveK and $16$ filters for the PPR10K, as mentioned above. 

We use the standard Adam optimizer~\cite{2015-kingma} to minimize the loss function in Equation~\ref{eq:3.2_4} and~\ref{eq:3.2_5}. The mini-batch size is $16$. We train models for 1,000,000 iterations with initial learning rate $l_{r}=2\times{10^{-4}}$. We use a cosine strategy to gradually decay the learning rate every 250,000 iterations. All experiments are conducted on NVIDIA Tesla V100 GPU.

In the following sections, we denote our main model in Section~\ref{sec:3.2} as {\em RSFNet-map}. Variations depicted in Section~\ref{sec:3.3} are denoted as {\em RSFNet-palette, RSFNet-saliency} and {\em RSFNet-semantic}.
\begin{table*}[t]
\centering
\small
\setlength{\tabcolsep}{0.5\tabcolsep}
\begin{tabular}{@{}ccccccccccc@{}}
\toprule
\multirow{2}{*}{Method} &
  \multicolumn{3}{c}{\begin{tabular}[c]{@{}c@{}}480p \\ zeroed with C's\end{tabular}} &
  \multicolumn{4}{c}{\begin{tabular}[c]{@{}c@{}}480p \\ zeroed as shot\end{tabular}} &
  \multicolumn{3}{c}{\begin{tabular}[c]{@{}c@{}}Full Resolution \\  zeroed as shot\end{tabular}} \\ \cmidrule(l){2-11} 
                 & PSNR$\uparrow$ & SSIM$\uparrow$ & $\triangle E_{ab}\downarrow$ & PSNR$\uparrow$ & SSIM$\uparrow$ &  $\triangle E_{ab}\downarrow$ & Runtime$\downarrow$ & PSNR$\uparrow$ & SSIM$\uparrow$ & Runtime$\downarrow$ \\ \midrule
UPE~\cite{DBLP:conf/cvpr/WangZFSZJ19}              & \ \ 21.88*    & \ \ 0.853*    &\ \ 10.80*   & /    & /      & /    &/     &/ &/ &/        \\
DPE~\cite{DBLP:conf/cvpr/ChenWKC18}                     & \ \ 23.75*    & \ \ 0.908*      &\ \ 9.34*   &/      &/      &/     &/   &/         &/ &/         \\
HDRNet~\cite{DBLP:journals/tog/GharbiCBHD17}   & \ \ 24.66*    &\ \ 0.915*   &\ \ 8.06* & /   & /      & /     &/ &/  &/      &/         \\
DeepLPF~\cite{DBLP:conf/cvpr/MoranMMPS20}        & \ \ 24.73*    &\ \ 0.916*  &\ \ 7.99*  & 23.38    &0.880      &10.03 &44.25         &23.40     &0.863       &1133.90         \\
CSRNet~\cite{DBLP:conf/eccv/HeLQD20}                 & \ \ 25.17*   &\ \ \textcolor{blue}{0.924}*   &\ \ 7.75*  &24.24    &0.910   &9.70   &\textcolor{blue}{3.49}         &23.04     &0.874    &80.6         \\
SA-3DLUT~\cite{DBLP:conf/iccv/Wang0PMWSY21}  &\ \ \textcolor{red}{25.50}*  &/  & /      &/     &/      &/     &/    &/       &/     & /        \\
3D-LUT~\cite{zeng2020lut}                                        &\ \ 25.29*      &\ \ 0.923*  &\ \ 7.55*   &/      &/   &/   &/         &/ &/ &/      \\
3D-LUT+AdaInt~\cite{yang2022adaint}                   &\ \ \textcolor{blue}{25.49}*      &\ \ \textcolor{red}{0.926}*   &\ \ \textcolor{blue}{7.47}*  &\textcolor{blue}{24.50}      &\textcolor{blue}{0.912}   &\textcolor{blue}{9.22}   &\textcolor{red}{1.59}         &\textcolor{blue}{24.24}       &0.857   &\textcolor{red}{1.80}         \\
\textcolor{violet}{Harmonizer}~\cite{Harmonizer}                                     &24.11       &0.904   &8.23   &23.23      &0.893  &10.14    &16.74         &22.57       &0.870   &27.98         \\ \midrule
\textcolor{violet}{RSFNet-map} &\textcolor{blue}{25.49}     &\textcolor{blue}{0.924}   &\textcolor{red}{7.23}   &\textcolor{red}{24.64}      &\textcolor{red}{0.915}   &\textcolor{red}{9.16}  &9.98         &\textcolor{red}{24.39}       &\textcolor{red}{0.894}   &12.35         \\
\textcolor{violet}{RSFNet-palette}    &25.01       &0.914   &7.62   &24.22    &0.911   &9.52   &19.76         &23.88       &0.888  &68.57         \\
\textcolor{violet}{RSFNet-saliency}   &24.78       &0.916    &7.86  &24.20      &0.912    &9.42  &14.09         &24.00       &0.890   &59.42         \\
\textcolor{violet}{RSFNet-semantic}   &24.76       &0.915    &7.77 & 24.19      &\textcolor{blue}{0.912}   &9.45   &24.07         &23.89       &\textcolor{blue}{0.891}  &71.18         \\
\textcolor{violet}{RSFNet-global}    &24.31      &0.911   &8.21  &23.43      &0.904    &10.16  &9.42         &23.26       &0.885    &\textcolor{blue}{10.99}         \\ \bottomrule
\end{tabular}%
\caption{Quantitative comparisons for retouching tasks on FiveK dataset~\cite{fivek}. Runtime is measured in miliseconds. ``*" means the result is adopted from the paper~\cite{yang2022adaint}. ``/" means the result is not available. Runtime shown in previous works are not adopted due to different hardware settings. Results with the first and second performance are colored as red and blue respectively. White-box methods are colored as violet. Full resolution images results are evaluated using Python, while others are evaluated on Matlab.}
\label{tab:4.4_1}
\end{table*}
\begin{table*}[t]
\centering
\small
\setlength{\tabcolsep}{1.0\tabcolsep}
\begin{tabular}{@{}ccccccc@{}}
\toprule
\multirow{2}{*}{Method} &
  \multicolumn{2}{c}{\begin{tabular}[c]{@{}c@{}}360p\\               expert a    \end{tabular}} &
  \multicolumn{2}{c}{\begin{tabular}[c]{@{}c@{}}360p\\               expert b    \end{tabular}} &
  \multicolumn{2}{c}{\begin{tabular}[c]{@{}c@{}}360p\\                expert c     \end{tabular}}  \\ \cmidrule(l){2-7} 
                  & PSNR$\uparrow$ & SSIM$\uparrow$ & PSNR$\uparrow$ & SSIM$\uparrow$ & PSNR$\uparrow$ & SSIM$\uparrow$ \\ \midrule
DeepLPF~\cite{DBLP:conf/cvpr/MoranMMPS20}            & 23.47    & 0.892    & 22.77   &0.875      &23.73      &0.896     \\
CSRNet~\cite{DBLP:conf/eccv/HeLQD20}           & 24.01    & 0.936    & 23.91    &0.938      &24.31      &0.931   \\
3D-LUT+AdaInt~\cite{yang2022adaint}    &\textcolor{red}{25.98}      &\textcolor{blue}{0.947}      &\textcolor{red}{24.91}      &0.936      &\textcolor{blue}{25.48}      &0.919      \\
\textcolor{violet}{Harmonizer}~\cite{Harmonizer}      &24.76      &0.929      &22.79      &0.885      &24.56      &0.902   \\ \midrule
\textcolor{violet}{RSFNet-map}  &\textcolor{blue}{25.58}      &\textcolor{red}{0.949}      &\textcolor{blue}{24.81}      &\textcolor{red}{0.945}      &\textcolor{red}{25.52}      &\textcolor{red}{0.939}       \\
\textcolor{violet}{RSFNet-saliency} &25.53      &0.946      &24.72      &\textcolor{blue}{0.944}      &25.11      &\textcolor{blue}{0.939}      \\ \bottomrule
\end{tabular}%
\caption{Quantitative comparisons for retouching tasks on PPR10K dataset~\cite{ppr10k}. All the models are trained on data without augmentations.White-box methods are colored as violet.}
\label{tab:4.4_2}
\end{table*}
\subsection{Ablation Studies}
\label{sec:4.3}
In this section, we evaluate the ability of our frameworks in different settings.
\begin{figure*}[t]
 \centering
   \includegraphics[width=1.0\linewidth]{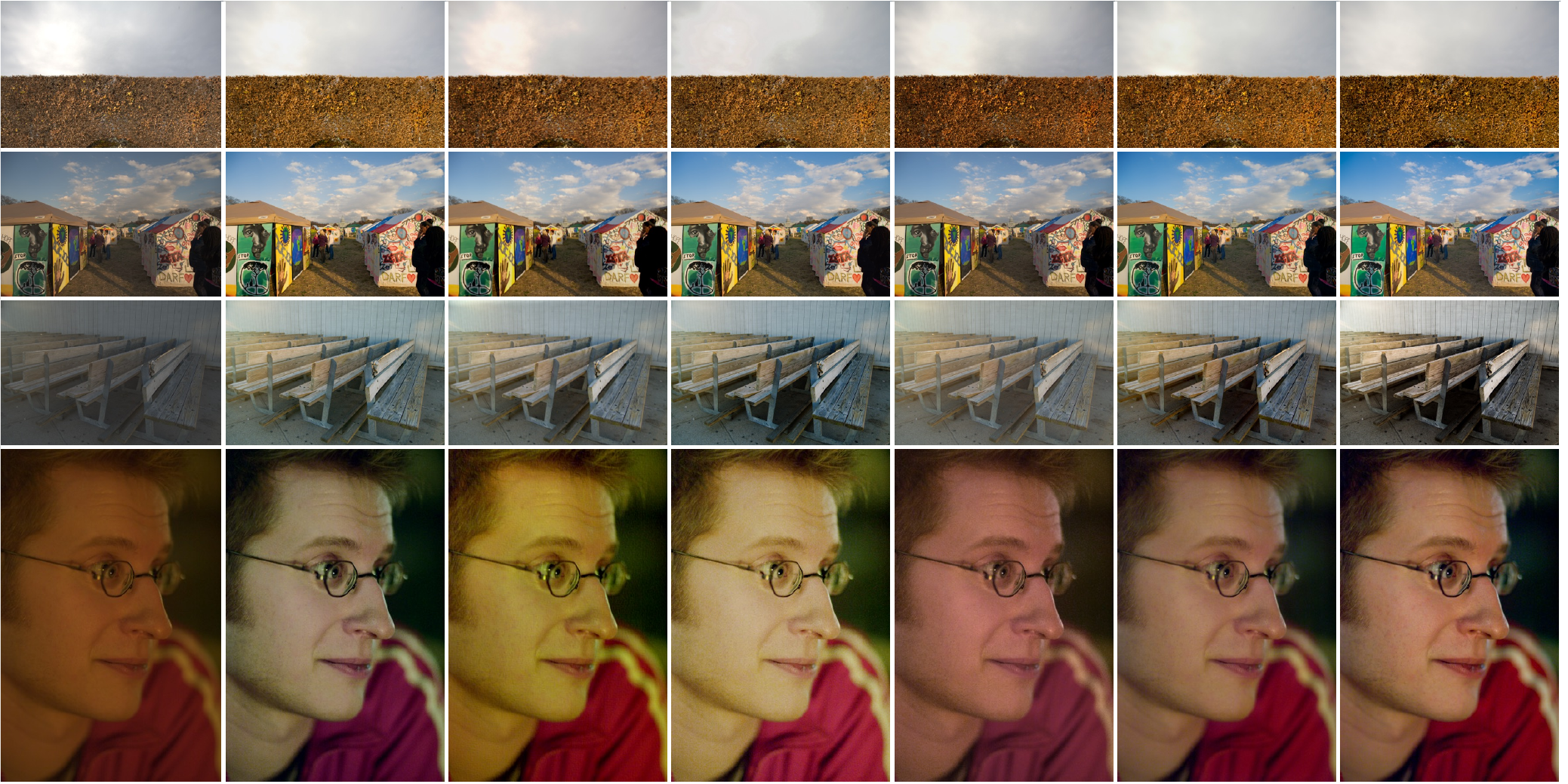}
     \small
\\
\begin{tabular}{p{2.05cm}<{\centering} p{2.05cm}<{\centering} p{2.05cm}<{\centering} p{2.05cm}<{\centering} p{2.05cm}<{\centering} p{2.05cm}<{\centering} p{2.05cm}<{\centering}}
input & CSRNet & DeepLPF & AdaInt & Harmonizer & RSFNet-map & GroundTruths
\end{tabular}%
   \caption{Qualitative comparison of selected methods. ``RSFNet-map" means our main model explained in Section~\ref{sec:3.2}}
   \label{fig:img_compare}
\end{figure*}
\begin{figure}
    \centering
    \includegraphics[width=0.45\textwidth]{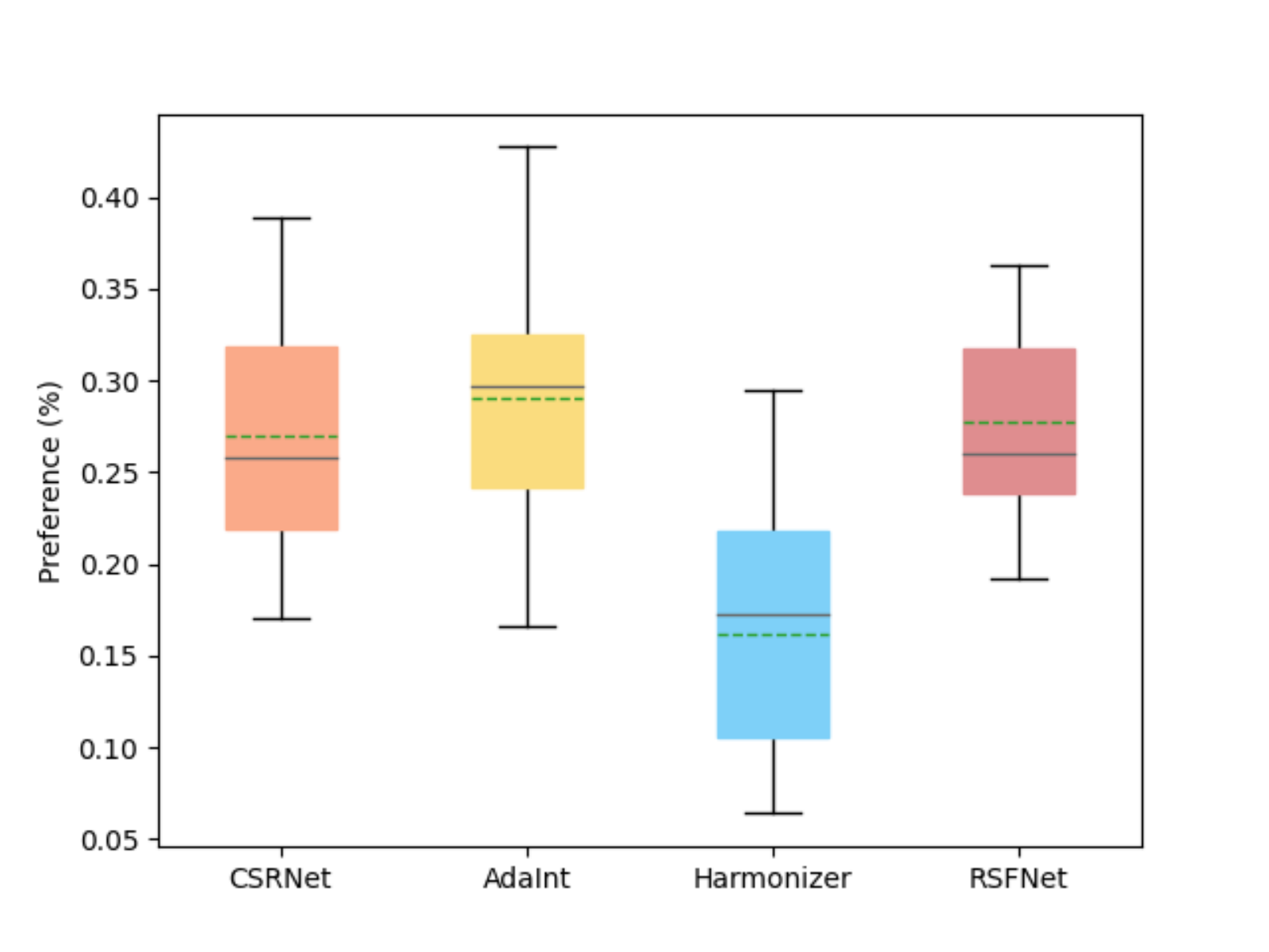}
    \caption{Boxplot of user study. The dashed green line and the solid black line inside the box are the mean and the median preference percentage respectively.}
    \label{fig:user_study}
\end{figure}

{\bf Comparative study of parallel and sequential approaches.}
We conduct experiments to compare parallel and sequential approaches under the same set of filter functions and backbone structures. For sequential implementations, we modify the network structure to mimic artistic workflows and resemble Harmonizer~\cite{Harmonizer}. We train the network using randomly shuffled 20 unique sequences of filter functions. The results are shown in Figure~\ref{fig:seq_par}. The parallel approach shows Pareto-optimality with near-maximum PSNR and a lower PSNR standard deviation (std). It also outperforms all sequential approaches in terms of SSIM and SSIM std. We find that altering the sequence of applying filters can lead to considerable changes in performance. Nearly half of these sequences fail to produce satisfactory results. Moreover, during training, the parallel approach converges faster than its sequential counterparts. In addition, the parallel method can be faster than the sequential one.

{\bf Training map generation and arguments regression separately.}
To train models with controlled region shapes, we experiment with using pre-trained PoolNet~\cite{liu2019simple} and adding an FPN neck followed by an argument regressor to the backbone instead of training from scratch. During training, we fix the weights of the backbone to keep the saliency mask generated by the network unchanged. The quantitative results are shown in Table~\ref{tab:4.3_1.2}, which indicates that there is no superiority over training from scratch. The reason for this might be that a network trained only on semantic data lacks sufficient information for retouching. Therefore, it is better to train the map generation and arguments regression tasks simultaneously.

{\bf Using Masks as Input.}
For training models with controlled region shapes, we can use masks concatenated with the image as input for our model. We train three models using three sets of masks as input, respectively. The results are shown in Table~\ref{tab:4.3_1.1}. The retouching results evaluated on PSNR and SSIM~\cite{1284395} are lower than in other settings. This may be because the ground truth masks are generated by off-the-shelf models, which may differ from the underlying real masks of the dataset. Masks are slightly modified by the map generator in the inference stage to achieve better retouching results. Therefore, simple concatenation of masks with the input image can lead to worse performance compared to our design of the variant of RSFNet, which generates masks and filter arguments simultaneously.

{\bf Trade Off Between Speed and Accuracy.}
As we decrease the number of downsample operations in the backbone, the feature size grows larger and time costs increase. However, quantitative results of retouching also become better, as shown in~\ref{tab:4.3_1.3}. Speed is measured in miliseconds. 

\begin{figure*}[t]
 \footnotesize
\begin{tabular}{rc}
\begin{tabular}[c]{@{}r@{}}   \\\\ input\end{tabular}        & \multirow{10}{*}{
   \includegraphics[width=0.8\linewidth]{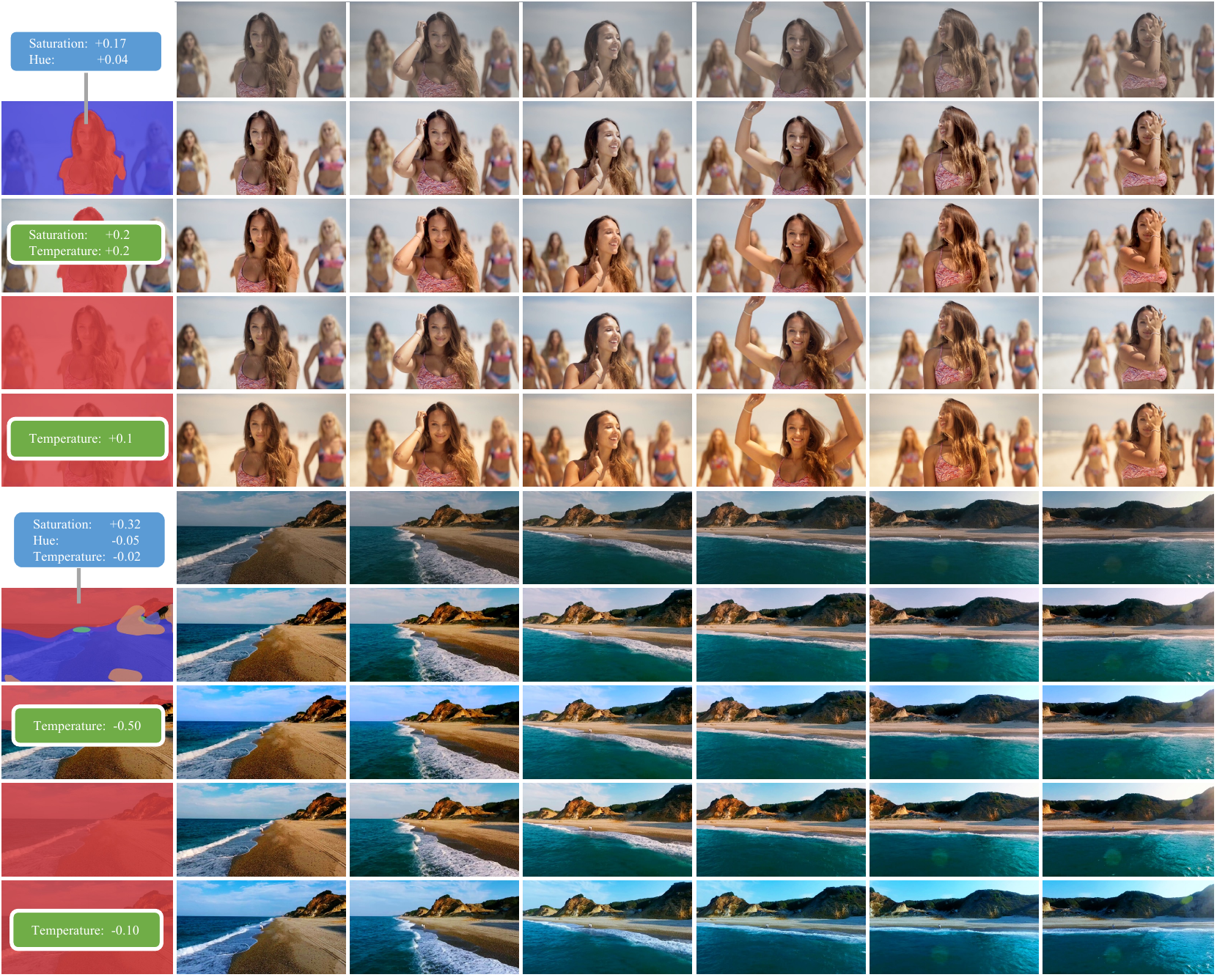}
} \\[35pt]
RSFNet-saliency                                                          &                     \\[15pt]
\begin{tabular}[c]{@{}r@{}}RSFNet-saliency\\ (Edit Version)\end{tabular} &                     \\[25pt]
Harmonizer                                                               &                     \\[15pt]
\begin{tabular}[c]{@{}r@{}}Harmonizer\\ (Edit Version)\end{tabular}      &                     \\[25pt]
input                                                                    &                     \\[23pt]
RSFNet-semantic                                                          &                     \\[15pt]
\begin{tabular}[c]{@{}r@{}}RSFNet-semantic\\ (Edit Version)\end{tabular} &                     \\[23pt]
Harmonizer                                                               &                     \\[17pt]
\begin{tabular}[c]{@{}r@{}}Harmonizer\\ (Edit Version)\end{tabular}      &                    
\end{tabular}%
\\\\
   \caption{Editable image retouching. ``RSFNet-saliency", ``RSFNet-semantic" means RSFNet trained with saliency masks and semantic masks respectively.} 
   \label{fig:video_edit}
\end{figure*}

\subsection{Comparison with State-of-the-Art}
\label{sec:4.4}
We compare our methods with the state-of-the-art for black-box and white-box image retouching tasks.
The selected methods are compared on PSNR, SSIM~\cite{1284395}, the $L_{2}$-distance in CIE LAB color space ($\triangle E_{ab}$) and the inference speed. We follow the practice in~\cite{yang2022adaint} to measure the GPU inference time on 100 images and report the average. Quantitative results are shown in Table~\ref{tab:4.4_1} and ~\ref{tab:4.4_2}. {\em RSFNet-map} refers to our main model in Section~\ref{sec:3.2}, {\em RSFNet-saliency},  {\em RSFNet-palette} and {\em RSFNet-semantic} refers to models trained with three sets of masks respectively in Section~\ref{sec:3.3}.
In addition, we evaluate a global retouching model without the map generator, denoted as {\em RSFNet-global}. Other models are trained using their official public codes and default configurations. All experiments are executed on an NVIDIA Tesla V100 GPU. 

Our proposed method, {\em RSFNet-map}, demonstrates its effectiveness in temperature correction and color enhancement, especially for the second set of inputs under more deteriorating shooting conditions. Although {\em RSFNet-map} achieves state-of-the-art results with negligible increase, while other RSFNet variations lag behind the state of the art, it is important to note that our white-box framework provides human-understandable ways for retouching, which makes it more convenient for users to edit and assess retouching results compared to other black-box methods. When compared with state-of-the-art white-box methods such as Harmonizer~\cite{Harmonizer}, all of our models exhibit better performance with faster speed. Furthermore, using 3D-LUTs to encode filter functions could potentially accelerate our models, a technique commonly implemented in traditional retouching tools. 
For PPR10K~\cite{ppr10k}, we also evaluate our methods using a random split setup. The results are provided in Appendix \ref{appendix:c}, consistently reinforcing the conclusions drawn from previous experiments. 

The qualitative results presented in Figure~\ref{fig:img_compare} demonstrate that our proposed method, {\em RSFNet-map}, exhibits superior performance in handling color transitions across regions, particularly in highlight areas, compared to 3D-LUT based methods such as AdaInt, as shown in the first row of the figure. Furthermore, we conduct a comparison with Harmonizer~\cite{Harmonizer} on editable retouching, as shown in Figure~\ref{fig:video_edit}. Our white-box framework offers more degrees of freedom than global-retouching manipulations to achieve region-specific retouching. For instance, when the temperature is adjusted globally using Harmonizer, the sky in the first image turns yellow. On the other hand, our {\em RSFNet-saliency} model can modify the temperature of the foreground girl while leaving other regions unaffected. In the second image, global temperature adjustment by Harmonizer turns the lake and mountains blue. In contrast, {\em RSFNet-semantic} only modifies the sky temperature, leaving the mountains and lake unchanged.

To further validate the effectiveness of our proposed framework, we conduct a user study to evaluate human preferences for RSFNet and other state-of-the-art methods, including CSRNet~\cite{DBLP:conf/eccv/HeLQD20}, AdaInt~\cite{yang2022adaint}, and Hormonizer~\cite{Harmonizer}. To form the test set, we randomly select $58$ images from the validation set of Adobe-MIT FiveK using the {\em input zeroed with expertC white balance} and {\em input zeroed as shot} settings, as well as an additional $20$ images downloaded from the internet. During the experiment, we display retouched images produced by all methods, including the original input, to $22$ participants. Participants are asked to select the best result from a randomly shuffled set of retouched images generated by all methods. We calculate the preferences of participants for each method and plot the results using a box plot, as shown in Figure~\ref{fig:user_study}.

Among the compared methods, AdaInt~\cite{yang2022adaint} achieves higher mean and median percentages. RSFNet exhibits more stable performance with a higher bottom percentage and the smallest standard variance percentage than the other methods. In comparison to the white-box method Harmonizer~\cite{Harmonizer}, RSFNet demonstrates superior user preference percentages. The boxplot illustrates the varying preferences of different users, highlighting the significance of editable capability. As RSFNet utilizes traditional color filters for retouching, it can be readily integrated into traditional retouching tools. Users can further enhance the visual appeal of the retouched results according to their own preferences through an interface.

{\bf Video Retouching.}
Our proposed RSFNet model also demonstrates applicability to video retouching, as shown in Figure~\ref{fig:video_edit}. The filter arguments remain constant within a single video clip for all frames, ensuring consistency throughout the retouching process. However, for RSFNet variations with controlled region shapes, such as {\em RSFNet-saliency}, the retouching consistency across frames relies on the consistency of region masks across frames. Although filter arguments can alleviate the inconsistency caused by masks, severe inconsistency problems in masks may still affect retouching results. With the assistance of a more robust tracking algorithm, these models can achieve better performance in editable video retouching, highlighting the potential of RSFNet in various retouching applications.

\section{Limitation and Conclusion}
We develop our framework under the assumption that all adjustments are done in one {\em layer}, which is then combined via linear summations of increments of all filters to obtain the final result. Therefore, although our model provides a white-box retouching framework that aligns with the intuition of human artists, it is unable to cover all adjustments that artists could conduct using professional software. However, we believe that if the cache data in the retouching process of artists is available, such as the region masks used for region-specific retouching, our model equipped with a more complicated structure, such as a cascaded structure in~\cite{hu2018exposure, Harmonizer}, could behave more similarly to a real human artist by learning from this data.

This paper introduces RSFNet, a white-box framework for image retouching that utilizes a divide-and-conquer strategy to generate region maps and human-understandable filter arguments. The resulting filtered images are combined using linear summations, allowing for a wider range of filter classes and achieving fine-grained enhancement with superior performance. A variation of RSFNet with controlled shape region masks is also proposed for user convenience. Extensive experiments are presented to demonstrate the effectiveness of RSFNet.
\\

\noindent \textbf{Acknowledgement}~This work was supported by Alibaba Group through Alibaba Innovative Research (AIR) Program and Alibaba-NTU Singapore Joint Research Institute (JRI), Nanyang Technological University, Singapore. We appreciate Mr. Wang Yuxi's efforts in generating additional experimental results during the rebuttal period.

{\small
\bibliographystyle{ieee_fullname}
\bibliography{egbib}
}


\begin{appendices}
\section{Filter Function}
\label{appendix:a}
The retouched result is represented as equation:
\begin{equation}
\begin{aligned}
  Y =  X + \sum_{m} \sum_{n} (F_{m,n}(\theta_{m,n}, X)-X)\odot M_{m},
  \end{aligned}
  \label{eq:1.1_1}
\end{equation}
For different filters, $F_{m,n}(\theta_{m,n}, X)$ has different expressions:
\begin{equation}
\begin{aligned}
 F_{contrast}(\theta, X)-X &= \theta(X - mean(X)) \\
 F_{saturation}(\theta, X)-X &= \theta(X - L(X)) \\
 F_{hue}(\theta, X_{c})-X_{c} & = \left\{
 \begin{aligned}
\alpha_{h} \theta X_{c} &, X_{c} \in \{X_{R}, X_{G}\}  \\
-\frac{1}{2} \alpha_{h} \theta X_{c} &, X_{c} = X_{B} \\
\end{aligned}
\right. \\
F_{temperature}(\theta, X_{c})-X_{c} &=  \left\{
 \begin{aligned}
\alpha_{t,1} \theta X_{c} &, X_{c} = X_{R}, \theta \geq 0  \\
\alpha_{t,2} \theta X_{c} &, X_{c} = X_{R}, \theta < 0 \\
0.0 &, X_{c} = X_{G}, \theta \geq 0 \\
\alpha_{t,3} \theta X_{c} &, X_{c} = X_{G}, \theta < 0 \\
\alpha_{t,4} \theta X_{c} &, X_{c} = X_{B}, \theta \geq 0  \\
\alpha_{t,5} \theta X_{c} &, X_{c} = X_{B}, \theta < 0 \\
\end{aligned}
\right. \\
F_{shadows}(\theta, X)-X  &= \theta (1 - X) \\
F_{midtones}(\theta, X)-X &= \theta (0.25-(X-0.5)^{2}) \\
F_{highlights}(\theta, X)-X &= \theta X \\
F_{shift}(\theta, X_{c})-X_{c} &=  \left\{
 \begin{aligned}
 \alpha_{s,1} &, X_{c} = X_{R} \\
 \alpha_{s,2} &, X_{c} = X_{G} \\
 \alpha_{s,3} &, X_{c} = X_{B} \\
 \end{aligned}
\right. \\
   \end{aligned}
  \label{eq:1.1_2}
\end{equation}
Where $mean(X)$ denotes the mean value of the entire image, while $L(X)$ represents the L channel of the image in the CIE LAB color space. Additionally, $X_{c} \in {X_{R}, X_{G}, X_{B}}$ denotes the RGB color channels of the image. The adjustment factor, denoted by $\alpha$, is a scalar that satisfies $\alpha > 0$. In our experiments, values of $\alpha$ are determined according to traditional color grading tools.

\section{Variation of RSFNet with Controlled Region Shape}
\label{appendix:b}
{\bf Ground Truths Mask Generation.}
We adopt the palette-based method proposed in~\cite{Chang_2015_PPR} to generate the main colors $C = {C_{1},..., C_{n}}$ of an image, along with distance maps from pixels to $N$ color centers. We obtain region masks by applying a Gaussian smoothing function to these distance maps, resulting in $M_{palette} = {M_{p1},...M_{pK_{p}}}$. We also predict saliency masks using pre-trained networks from~\cite{liu2019simple}, yielding $M_{saliency}={M_{sal1},M_{sal2}}$.

Since previous works on panoptic segmentation split objects into more than one hundred classes, which is redundant for our task, we aim to identify the most significant pixel groupings. To accomplish this, we follow the practice in~\cite{caron2021emerging,van2022discovering} and train a self-attentioned network with pairwise retouching data. First, we predict semantic masks using the networks presented in~\cite{cheng2021mask2former}. We then apply a clustering algorithm (e.g., K-means~\cite{lloyd1982least}) to the output features of the self-attentioned network with masked images as input. Masks assigned with the same cluster index are merged, resulting in $M_{semantics} = {M_{sem1},...M_{semK_{s}}}$. For each of the three sets of masks, we train a separate model with the corresponding output channel numbers. The entire process is illustrated in Figure~\ref{fig:3.3_1}.
~\\

{\bf Differentiable Adaptive Smooth Kernel.}
To ensure smooth transition across mask edges, we have incorporated a differentiable adaptive smooth kernel module into our main network. We fix the Gaussian smooth kernel to a suitable size $\sigma_{max}$, such as $51 \times 51$ for the original input with a resolution of $256 \times 256$. The standard variance of the kernel is a learnable parameter, which adapts to the inputs.

\section{Additional Results}
\label{appendix:c}
{\bf Quantitative Results.}
We valuate our methods using a random split setup for PPR10K~\cite{ppr10k}. The results are demonstrated in Table~\ref{tab:random}. For more implementation details, please refer to our codebase at {\small \url{https://github.com/Vicky0522/RSFNet}}.
\begin{table*}[t]
\centering
\small
\setlength{\tabcolsep}{1.0\tabcolsep}
\begin{tabular}{@{}ccccccc@{}}
\toprule
\multirow{2}{*}{Method} &
  \multicolumn{2}{c}{\begin{tabular}[c]{@{}c@{}}360p\\               expert a    \end{tabular}} &
  \multicolumn{2}{c}{\begin{tabular}[c]{@{}c@{}}360p\\               expert b    \end{tabular}} &
  \multicolumn{2}{c}{\begin{tabular}[c]{@{}c@{}}360p\\                expert c     \end{tabular}}  \\ \cmidrule(l){2-7} 
                  & PSNR$\uparrow$ & SSIM$\uparrow$ & PSNR$\uparrow$ & SSIM$\uparrow$ & PSNR$\uparrow$ & SSIM$\uparrow$ \\ \midrule
DeepLPF~\cite{DBLP:conf/cvpr/MoranMMPS20}            & 24.97    & 0.939    & 24.33   &0.930      &24.65      &0.926     \\
CSRNet~\cite{DBLP:conf/eccv/HeLQD20}           & 24.38    & 0.938    & 24.41    &0.940      &24.53      &0.931   \\
3D-LUT+AdaInt~\cite{yang2022adaint}    &\textcolor{red}{27.31}      &\textcolor{blue}{0.954}      &\textcolor{red}{26.62}      &0.945      &\textcolor{blue}{26.67}      &0.929      \\
\textcolor{violet}{Harmonizer}~\cite{Harmonizer}      &25.02      &0.916      &23.84      &0.895      &25.22      &0.920    \\ \midrule
\textcolor{violet}{RSFNet-map}  &\textcolor{blue}{27.25}      &\textcolor{red}{0.956}      &\textcolor{blue}{26.61}      &\textcolor{red}{0.954}      &\textcolor{red}{26.76}      &\textcolor{red}{0.945}       \\
\textcolor{violet}{RSFNet-saliency} &25.98      &0.946      &25.87      &\textcolor{blue}{0.948}      &25.88      &\textcolor{blue}{0.937}      \\ \bottomrule
\end{tabular}%
\caption{Quantitative comparisons for retouching tasks on PPR10K dataset~\cite{ppr10k}. All the models are trained on data without augmentations.White-box methods are colored as violet.}
\label{tab:random}
\end{table*}

\begin{figure*}
    \centering
    \includegraphics[width=0.99\textwidth]{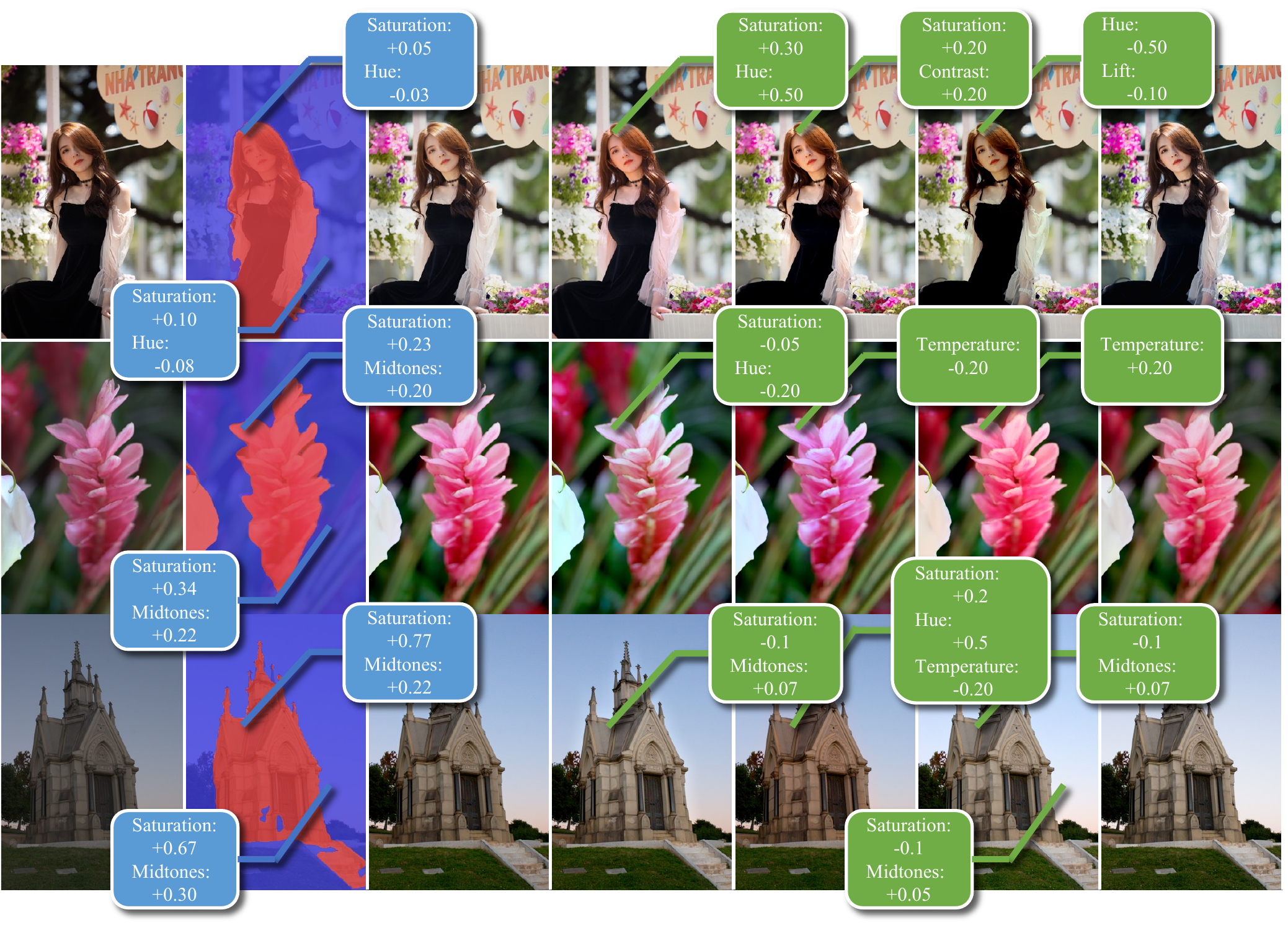}
     \resizebox{\textwidth}{!}{%
  \footnotesize
   \begin{tabular}{p{2.05cm}<{\centering} p{2.05cm}<{\centering} p{2.05cm}<{\centering} p{2.05cm}<{\centering} p{2.05cm}<{\centering} p{2.05cm}<{\centering} p{2.05cm}<{\centering}}
input & Result(Args \& Masks) & Result(Image) & Edit Version 1. & Edit Version 2. & Edit Version 3. & GroundTruths
\end{tabular}%
}
    \caption{Editable white-box retouching. Arguments and masks generated by RSFNet-saliency trained with saliency masks are shown in the second column. Retouched result is shown in the third column. Three versions of adjustments conducted on the retouched results are shown in the three columns on the right. Ground truths is shown in the right-most column. Numbers in green boxes indicate relative variation.}
    \label{fig:sal}
\end{figure*}

\begin{figure*}
    \centering
    \includegraphics[width=0.99\textwidth]{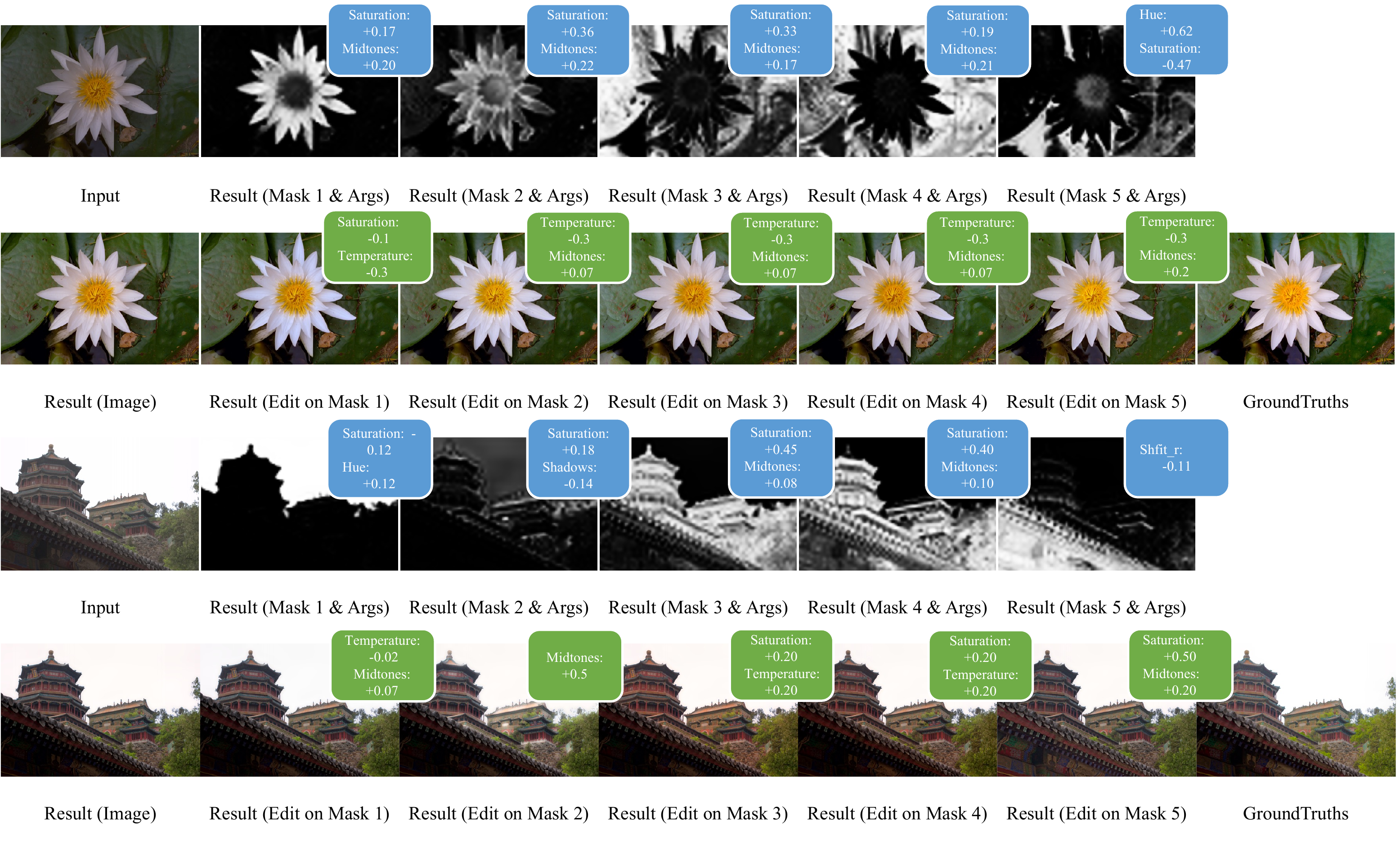}
    \caption{Editable white-box retouching. Arguments and masks generated by RSFNet-palette trained with palette-based masks are shown in the first row. Only two of the most significant arguments are presented. Retouched result is shown in the first column of the second row. Five versions of adjustments conducted on the retouched results and corresponding masks are shown in the rest columns of the second row. Ground truths is shown in the right-most column. Numbers in green boxes indicate relative variation.}
    \label{fig:pal}
\end{figure*}

{\bf Qualitative Results.}
We present more results of RSFNet-saliency, RSFNet-palette and RSFNet-map in Figure~\ref{fig:sal} and~\ref{fig:pal}, including generated masks and corresponding filter arguments. 

\end{appendices}

\end{document}